# AssemPlanner: A Multi-Agent Based Task Planning Framework for Flexible Assembly System

**Chenhao Zhang[a,†], Chaoran Zhang[a,†], Zhaobo Xu[a], YongboYang[a], Pingfa Feng[a,b], Long Zeng[a,*]**

a Tsinghua Shenzhen International Graduate School, Tsinghua University, Shenzhen 518055, China

b Department of Mechanical Engineering, Tsinghua University, Beijing 100084, China

* Corresponding author. E-mail address: zenglong@sz.tsinghua.edu.cn

† These authors equally contributed to this work.

**Abstract:**

In flexible assembly systems, existing task planning methods require a time-consuming configuration process by multiple experts to establish a production line for a new product. To address this challenge, we propose a multi-agent based task planning framework for flexible assembly systems, denoted as AssemPlanner. It takes tasks described in natural language as input, which are then converted into actionable sequential production operations. It comprises several specialized agents, including SchedAgent , KnowledgeAgent, LineBalanceAgent, and a scene graph. Within the proposed framework, SchedAgent serves as the central reasoning engine. Departing from traditional static pipelines, AssemPlanner utilizes a ReAct-based SchedAgent to adaptively adjust actions via multi-agent feedback. By observing the feedback from KnowledgeAgent, LineBalanceAgent, and the scene graph, it autonomously resolves complex industrial process constraints. In experiments, KnowledgeAgent achieves near-perfect accuracy on single-hop questions and 98.8% accuracy on multi-hop reasoning, effectively addressing the challenges of logical consistency and hallucination suppression in long-chain industrial reasoning; LineBalanceAgent achieves performance comparable to established scheduling methods such as Deep Q-learning (DQN) and meta-heuristic algorithms such as Artificial Bee Colony (ABC) and Monarch Butterfly Optimization (MBO). Significantly, it eliminates the need for expert-level parameter tuning and complex reward function design by enabling rapid task adaptation solely through textual prompt modifications; SchedAgent achieves over 68% accuracy in task planning and above 96% in subtask decomposition. In addition, a real-world case study of pressure valve assembly demonstrates that AssemPlanner significantly enhances deployment efficiency. To facilitate reproducibility, all code and datasets are released at https://github.com/chz332/Assemplanner.

# 1 Introduction

In flexible assembly systems [1] , establishing a new production line for a new product often necessitates repeated task decomposition, process design, and resource configuration conducted by multiple domain experts. This process is inherently time-consuming and labor-intensive, thereby limiting the responsiveness to multi-variety and small-batch manufacturing demands. Large Language Model (LLM) based agent systems in diverse cognitive tasks [2, 3], we investigate a question whether a similar multi-agent paradigm can be applied to task planning in flexible assembly. Specifically, we address the following question: Can a multi-agent based task planning framework enable autonomous understanding, reasoning, and coordination in assembly processes? The realization of such a system could significantly mitigate human intervention, expedite production line reconfiguration, and bolster adaptability across diverse product types and manufacturing contexts.

Existing assembly task planning methods be broadly classified into three categories. Knowledge-based methods [4] leverage knowledge graphs and Domain-Specific Language (DSL) to formalize and reason about complex process logic. However, in flexible assembly environments, these methods suffer from a lack of agility, as updating knowledge bases often requires manual intervention, leading to low cross-product generalizability. Learning-based methods [5] encompassing deep reinforcement learning and imitation learning, reduce the reliance on hard-coded rules but are typically data-hungry and require extensive retraining when faced with dynamic line reconfiguration. This makes rapid adaptation to novel constraints computationally expensive. Recently, LLM based methods [6] have demonstrated impressive semantic reasoning capabilities for high-level plan generation. Yet, these models often function in an open-loop fashion, lacking deep integration with domain-specific knowledge and low-level scheduling algorithms. Consequently, they struggle to generate executable action sequences that satisfy strict industrial constraints [7-9]. Crucially, none of these existing paradigms achieve a unified orchestration of semantic reasoning, domain constraints, and execution, hindering the realization of a truly adaptive, closed-loop control system.

To address the challenges of task planning of flexible assembly, we propose AssemPlanner, a novel multi-agent collaborative task planning framework designed for Embodied Intelligent Industrial Robot (EIIR) based physical reconfigurable flexible assembly systems [1, 10-13]. AssemPlanner represents a pioneering effort in implementing a comprehensive, natural-language-driven architecture for end-to-end assembly task planning.

Building upon our previous research [11], the framework adopts a hierarchical coordination architecture centered on the SchedAgent, which acts as the primary reasoning and decision-making hub. Utilizing a "Reason – Action – Observe" cycle, SchedAgent does not merely follow a static sequence, instead, it orchestrates dynamic negotiation by synthesizing real-time feedback from the KnowledgeAgent, LineBalanceAgent, and the scene graph. This closed-loop interaction allows the system to autonomously detect and resolve planning conflicts such as cycle time (CT) violations or resource dependencies, by adaptively updating its internal state and re-invoking specialized agents until all industrial constraints are satisfied. Specifically, the KnowledgeAgent employs a two-layer retrieval paradigm that synergizes semantic matching with topological graph expansion to ensure logical completeness and suppress hallucinations [14-16]. By enforcing structural constraints, this mechanism suppresses hallucinations and satisfies the reliability requirements of industrial assembly. The LineBalanceAgent optimizes process allocation via a natural-language-driven self-reflective strategy [17], while the scene graph maintains a structured spatial-resource representation. Following the ReAct-based reasoning paradigm [18], the SchedAgent progressively refines high-level task descriptions into actionable instructions, dynamically invoking these specialized agents as tools to ensure information integrity and constraint compliance. This orchestrated cooperation enables the autonomous transformation of vague requirements into optimized, executable assembly plans for flexible assembly.

Through a multi-level experimental validation, the AssemPlanner framework demonstrates superior overall performance in flexible assembly task planning, maintaining robust adaptability across cross-product generalization, real-time resource scheduling, and industrial line execution. Under two-shot setting, the framework achieves over 68% accuracy in full task planning and exceeds 96% accuracy in subtask decomposition, thereby verifying its end-to-end, closed-loop capability from natural-language instruction to executable plans. To evaluate the specific contributions of individual modules, core agents were independently tested across diverse subtask scenarios. The KnowledgeAgent reached 100% accuracy in single-hop industrial QA tasks and 98.8% in multi-hop reasoning, proving its capacity to provide precise, context-consistent domain knowledge for planning. Meanwhile, the LineBalanceAgent delivered performance competitive with DQN and meta-heuristic algorithms while replacing parameter tuning with natural-language interaction. This prompt-driven reflective optimization facilitates rapid task adaptation through textual instructions, removing the necessity for manual reward functions or complex encoding rules. Validation on a physical reconfigurable assembly line using a pressure valve assembly confirms that the multi-agent system successfully establishes a closed loop from natural-language input to robotic execution. Furthermore, the modular architecture of AssemPlanner supports independent deployment: the KnowledgeAgent can function as a standalone industrial reasoning service; the LineBalanceAgent can integrate into existing production schedulers; and the SchedAgent can generate executable subtask lists from external constraints. This versatile modularity enables the framework to operate either as a unified system or as decoupled components adaptable to diverse manufacturing environments.

The main contributions of this paper are as follows:

**(1) AssemPlanner (A Hierarchical Multi-Agent Coordination Framework):** We propose a unified framework that orchestrates multi-agent collaboration, LLM-based semantic reasoning, and industrial task planning. Unlike traditional static pipelines, AssemPlanner implements a closed-loop feedback-rectification mechanism, enabling the autonomous transformation of high-level natural language instructions into optimized, executable assembly sequences.

**(2) A Prompt-Driven Reflective Optimization Paradigm for Line Balancing:** We introduce a novel, natural-language-driven self-reflection strategy that achieves line balancing performance parity with established DQN and meta-heuristic baselines (e.g., ABC, MBO). By shifting the optimization paradigm from expert-intensive reward engineering to intuitive linguistic interaction, this approach significantly reduces deployment overhead.

**(3) A Symbolic-Grounding Mechanism via KG-enhanced RAG:** To ensure industrial-grade reliability, we develop a Knowledge Graph (KG) enhanced Retrieval-Augmented Generation (RAG) mechanism. By synergizing semantic similarity with topological structural, this mechanism provides rigorous symbolic grounding for LLM reasoning, effectively mitigating hallucinations and ensuring that all generated sub-tasks remain strictly consistent with intricate industrial process constraints.

# 2 Related Work

## 2.1 Industrial task planning

Industrial task planning is defined as the generation of sequential executable operations under stringent requirements and constraints. Unlike general-purpose task planning, industrial environments necessitate zero-tolerance for error, as planning inaccuracies can lead to catastrophic hardware failure or production downtime. For instance, while the sequence of tableware placement in a domestic setting is flexible, part positioning and process sequences in industrial assembly must strictly adhere to engineering specifications. This rigidity represents the primary challenge in adapting general planning methods to industrial contexts. Currently, existing assembly task planning methods can be broadly categorized into three paradigms:

**Knowledge and Skill-based Methods:** These have been extensively studied and can be further divided into Knowledge-graph based [19] and DSL based approaches. Merdan et al. and Hoebert et al. [20, 21] applied ontology-based knowledge frameworks to industrial robots. It reduces traditional programming complexity and configuration cost. Jiang et al. [22] combined digital twins with knowledge graph for assembly knowledge management, and Zhou et al. [23] built process knowledge graph for assembly plan generation and evaluation, while Qin and Lu [24] proposed a dynamic manufacturing semantic representation for adaptive control. DSL-based solutions, particularly those utilizing PDDL [25], standardize complex planning tasks. Kootbally et al. [26] integrated OWL knowledge with PDDL, and Rogalla et al. [27] employed PDDL to model discrete manufacturing processes. Despite their logical rigor, these methods suffer from rigid knowledge updates, heavy reliance on predefined rules, and limited adaptability in highly dynamic scenarios.

**Learning-based Methods:** Driven by advancements in deep learning, reinforcement learning (RL), and imitation learning [28], these methods extract planning logic directly from large datasets to reduce manual intervention. Deep learning has excelled in robotic grasping and human-robot collaboration [29]. Reinforcement learning is suitable for autonomous decision-making and data-scarce scenarios. For instance, Fan et al. [30] optimized workshop logistics navigation using deep reinforcement learning, and Jiang et al. [31] integrated knowledge graphs with reinforcement learning for assembly sequence planning. Imitation learning accelerates model training through expert demonstrations and has been applied in assembly tasks [32, 33]. Although these approaches mitigate the reliance on human-defined rules, they remain data-intensive and computationally expensive when generalized to new products.

**LLM-based Methods:** Large Language Models (LLMs) have been introduced to the industrial domain to leverage their natural language understanding for processing technical manuals and user feedback [6]. Tanaka and Katsura [34] achieved voice-controlled polishing robots; Wang et al. [35] combined LLM with visual navigation for manufacturing path planning; Fakih et al. [36] proposed the LLM4PLC framework to accelerate PLC programming; Fan et al. [37] and Gan et al. [38] applied LLM to multi-object rearrangement and task planning. Nevertheless, LLMs currently serve primarily as auxiliary modules for information interpretation rather than as the autonomous reasoning core of task planning.

In summary, existing methods face a significant bottleneck in the reconfiguration of flexible assembly systems. Knowledge and Skill-based methods lack agility in updates; learning-based methods struggle with cross-product generalization; and current LLM-based approaches still merely treat the large model as an auxiliary module, lacking deep integration with industrial constraints and production scheduling.

## 2.2 Multi-agent system

A Multi-Agent System (MAS) is an intelligent architecture comprising multiple autonomous entities with heterogeneous capabilities, designed to tackle complex tasks through collaboration, structured dialogue, and strategic interaction [39, 40]. Driven by the recent breakthroughs in LLM, MAS architectures have exhibited transformative potential across diverse domains, including mathematical problem-solving, software engineering, and business intelligence. Representative studies include: MetaGPT [2], which integrates standardized operating procedures into multi-agent systems to enhance collaboration efficiency; ChatDev [41] and AutoGen [3], which leverage communicative chains and interactive mechanisms to streamline complex software development; and HyperAgent [42] and Magentic-One [43], which demonstrate general-purpose capabilities in task decomposition and master-expert coordination. Furthermore, LLM+P [44] integrates classical symbolic planners with MAS to ensure the generation of reliable, optimal plans.

In the industrial sector, the collaborative and reasoning strengths of MAS provide a robust foundation for intelligent manufacturing, production scheduling, and predictive maintenance. ChatCNC [45] introduced a conversational monitoring system that utilizes RAG to allow operators to query real-time manufacturing data through natural language, significantly improving human-machine interaction. Gautam et al. [46] developed an Industrial IoT framework that bridges multi-agent LLMs with digital twins for unified data acquisition and real-time fault diagnosis. Similarly, Wang [47] introduced the Multi-Agent

Scheduling Chain framework, which employs semantic understanding and dynamic algorithm selection for flexible job-shop scheduling and real-time rescheduling, improving efficiency and adaptability. Sun et al. [48] integrated LLMs with digital twin technology to perform explainable, traceable, and adaptive fault diagnosis and decision-making through semantic time-series data and multi-agent collaborative reasoning. Liu et al. [49] further proposed an augmented framework combining knowledge-enhanced LLM and AR-guided visual assistance for machine tool fault reasoning and maintenance recommendation. By constructing fault scene graph and enabling multi-agent collaboration, their approach achieved high-precision fault localization, interpretable reasoning, and efficient repair operations.

In summary, while the aforementioned studies underscore the potential of MAS in industrial monitoring and scheduling, a critical gap remains in achieving autonomous, constraint-aware task planning for high-precision assembly. To address this, we propose AssemPlanner, a multi-agent framework specifically engineered to bridge the gap between high-level semantic intent and low-level execution logic. By facilitating iterative negotiation and dynamic feedback among the SchedAgent, KnowledgeAgent, and LineBalanceAgent, our framework enables the autonomous resolution of conflicting process and constraints via natural-language interaction. This modular and negotiation centric design not only ensures rigorous logical grounding but also allows for rapid, zero-shot adaptation to diverse industrial scenarios, establishing a new paradigm for flexible assembly.

## 3 Methods

This section details the system architecture of AssemPlanner and provides an in-depth description of its three constituent agents and their collaborative mechanisms.

### 3.1 Overview of AssemPlanner

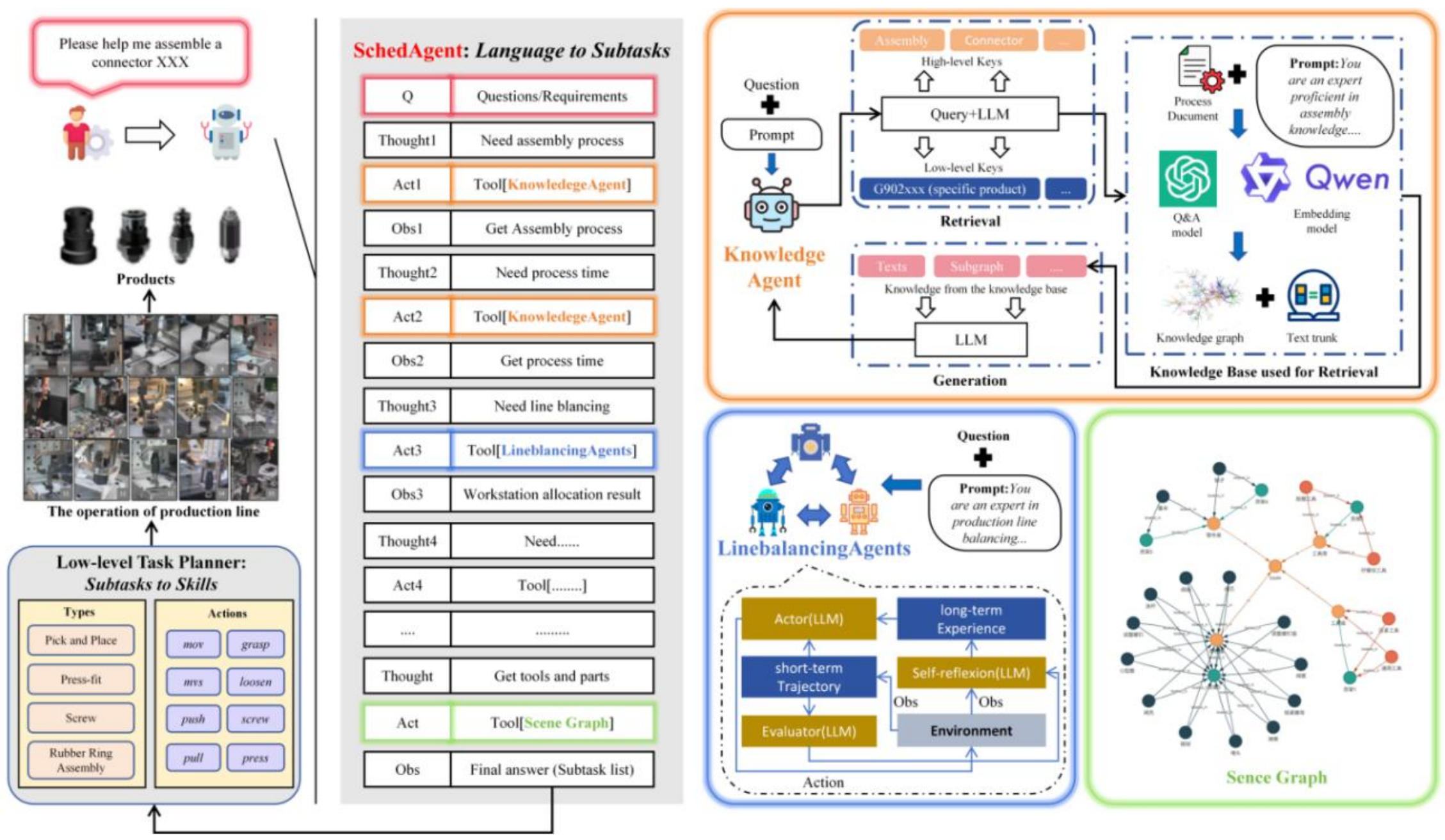

Fig.1 Framework of AssemPlanner

Within our previously proposed EIIR framework [11], the task planning and scheduling module functions as the cognitive hub bridging high-level intent and physical execution. The core component, AssemPlanner (Fig.1), transforms natural-language task descriptions into optimized, executable assembly sequences through multi-agent orchestration.

AssemPlanner adopts a hierarchical multi-agent architecture characterized by an ordered cooperation mechanism. It comprises three specialized agents: SchedAgent, KnowledgeAgent, and LineBalanceAgent, which facilitate task reasoning, knowledge retrieval, and line balancing optimization. The SchedAgent serves as the central reasoning engine. It receives a natural-language task $T_{NL}$ and initiates a reactive planning process. The overall objective is formalized as:

$$O = (T_{NL}, A_K, A_B, A_S, G) \tag{1}$$

where $T_{NL}$ denotes the user's natural-language input; $A_K$, $A_B$, $A_S$ represent the three core agents; $G$ is the scene graph providing environment and resource constraints; and $O$ denotes the final output assembly action sequence.

During the planning phase, SchedAgent follows the ReAct reasoning paradigm,

continuously perceiving feedback from other agents to adaptively refine the next reasoning step. Departing from static pipelines, SchedAgent functions as an autonomous coordinator that invokes KnowledgeAgent, LineBalanceAgent, and the scene graph to generate answers that comply with industrial constraints. Specifically, KnowledgeAgent utilizes a KG-enhanced RAG mechanism to provide precise industrial logic, while LineBalanceAgent performs self-reflective optimization to ensure CT compliance. The scene graph provides information about resources in the environment. In this manner, it progressively decomposes abstract tasks into structured subtasks such as material picking, loading, and assembly that are both logically sound and physically feasible.

Finally, the Low-level Task Planner maps these subtasks to specific robotic skills. By coupling natural-language reasoning with orchestrated agent cooperation, the system ensures a seamless, adaptive translation from user intent to concrete robotic actions across heterogeneous manufacturing scenarios.

In the following subsections, the approaches to construct the three agents are detailed.

## 3.2 SchedAgent

SchedAgent serves as the central reasoning engine of AssemPlanner, operating under an enhanced ReAct (Reason–Action–Observe) paradigm. Departing from conventional modular pipelines that follow rigid, predefined sequences, SchedAgent establishes a dynamic negotiation closed-loop. Rather than acting as a passive tool invoker, it functions as a "Cognitive Decision Hub" that adaptively adjusts reasoning paths in real-time based on environmental feedback and specialized constraints from other agents. In each iteration, SchedAgent maintains a dynamic working memory and achieves task objectives through the following cyclical steps:

**Reasoning:** Synthesizes the current global state to identify logical inconsistencies among subtasks or detect missing constraint information.

**Action:** Proactively generates interrogation requests targeting specialized agents or the environment to fill information gaps.

**Observation:** Summarize the effects brought by the current action and guide the next step of action.

SchedAgent is equipped with autonomous conflict identification: for instance, if LineBalanceAgent reports an infeasible production CT, SchedAgent initiates heuristic backtracking to trigger a re-planning loop, ensuring all constraints are reconciled.

Once all informational requirements are met, SchedAgent fuses domain knowledge, constraints, and resource information in the environment to produce a validated task decomposition and resource allocation plan. Formally, the decision making process is abstracted as:

$$T_{t+1} = f(A_s^t, M_K^t, M_B^t, G) \quad (2)$$

$T^{t+1}$ represents the subtask list generated at time $t+1$; $A_s^t$ denotes the internal state of SchedAgent at time $t$; $M_K^t$ represents messages from the KnowledgeAgent $A_K$ at time $t$; $M_B^t$ represents messages from the LineBalancingAgent $A_B$, such as process allocation plans and CT constraints; $G$ denotes the scene graph, which provides semantic information about environment, resources, and constraints.

The function $f(\bullet)$ corresponds to the ReAct reasoning process: First, reason semantic parsing of the natural-language task; Then, action dynamically calls $A_K$、$A_B$ and G and to obtain required information; Finally, observing integrates external feedback and updates the state until a constraint compliant task decomposition plan $T_{t+1}$ is generated. This formula signifies that the final plan is not generated in a single pass but converges through inter-agent negotiation. The complete reasoning process of SchedAgent is shown in Algorithm 1. And figure 2 shows how SchedAgent handles conflicts by taking the example that the CT constraint cannot be met.

Algorithm1: AssemPlanner ("Assemble Valve")

**Input:** Natural language query $I$ = "Assemble one valve";
Toolset $=\{A_K,\ A_B,\ G\}$;
Scene graph G (environment information including rooms, shelves, resources).
**Output:** Final task list $T$ containing assembly steps with tool/part allocation.
1: Initialize time step $t \leftarrow 0$;
2: Initialize context $c_0 \leftarrow \emptyset$;
3: Receive natural language query $I$;
4: Update context $c_0 \leftarrow c_0 \cup \{I\}$;
5: **while** Final Answer not produced **do**
6: $t \leftarrow t+1$;
7: /* **Thought:** Reason about what information is required next */
8: **if** Assembly process unknown then /* **Action:** Call $A_K$ to find the assembly process */
9: $a_t \leftarrow$ InvokeTool($A_K$, "Retrieve assembly procedure for valve");
10: $o_t \leftarrow$ GetObservation($a_t$); /* **Observation:** Assembly process has been obtained */
11: Update context;
12: **else if** Task times unknown **then** /* **Action:** Call $A_K$ to find the assembly times */
13: $a_t \leftarrow$ InvokeTool($A_K$, "Retrieve execution time of each step");
14: $o_t \leftarrow$ GetObservation($a_t$); /* **Observation:** Assembly times has been obtained */
15: Update context $c_t \leftarrow c_{t-1} \cup \{a_t, o_t\}$;
16: **else if** Line balancing not solved **then** /* **Action:** Call $A_B$ to balance the production line */
17: $a_t \leftarrow$ InvokeTool($A_B$, {steps, times});
18: $o_t \leftarrow$ GetObservatio$a_t$n(); /* **Observation:** Production line balancing has been completed */
19: Update context $c_t \leftarrow c_{t-1} \cup \{a_t, o_t\}$;
20: **else if** Tools and parts unknown **then** /* **Action:** Call $A_K$ to obtain part and tool information */
21: $a_t \leftarrow$ InvokeTool($A_K$, "Retrieve required tools and parts for each step");
22: $o_t \leftarrow$ GetObservation($a_t$); /* **Observation:** Part and tool information has been obtained */
23: Update context $c_t \leftarrow c_{t-1} \cup \{a_t, o_t\}$;
24: **else if** Locations of resources unknown **then** /* **Action:** Call $G$ get the resources of the current scene */
25: $a_t \leftarrow$ InvokeTool($G$, "Retrieve location of tools and parts");

| | |
|---|---|
| 26: | $o_t$ ← GetObservation($a_t$); /* **Observation:** The resources of the current scene have been obtained */ |
| 27: | Update context $c_t \leftarrow c_{t-1} \cup \{a_t, o_t\}$; |
| 28: | **else** |
| 29: | /* **Thought:** Integrate knowledge, line balancing, and scene graph */ |
| 30: | $T$ ← GenerateFinalTaskList($c_t$, $G$); |
| 31: | Output $T$ as Final Answer; |
| 32: | return $T$; |
| 33: | **end if** |
| 34: | **end while** |

In summary, this multi-agent collaborative mechanism marks a paradigm shift in industrial task planning, transitioning from rigid, linear pipelines to a dynamic, iterative negotiation loop. By empowering the SchedAgent to adaptively synthesize formalized domain knowledge, spatial constraints from the scene graph, and line balancing optimization feedback, the system manifests a form of collaborative intelligence.

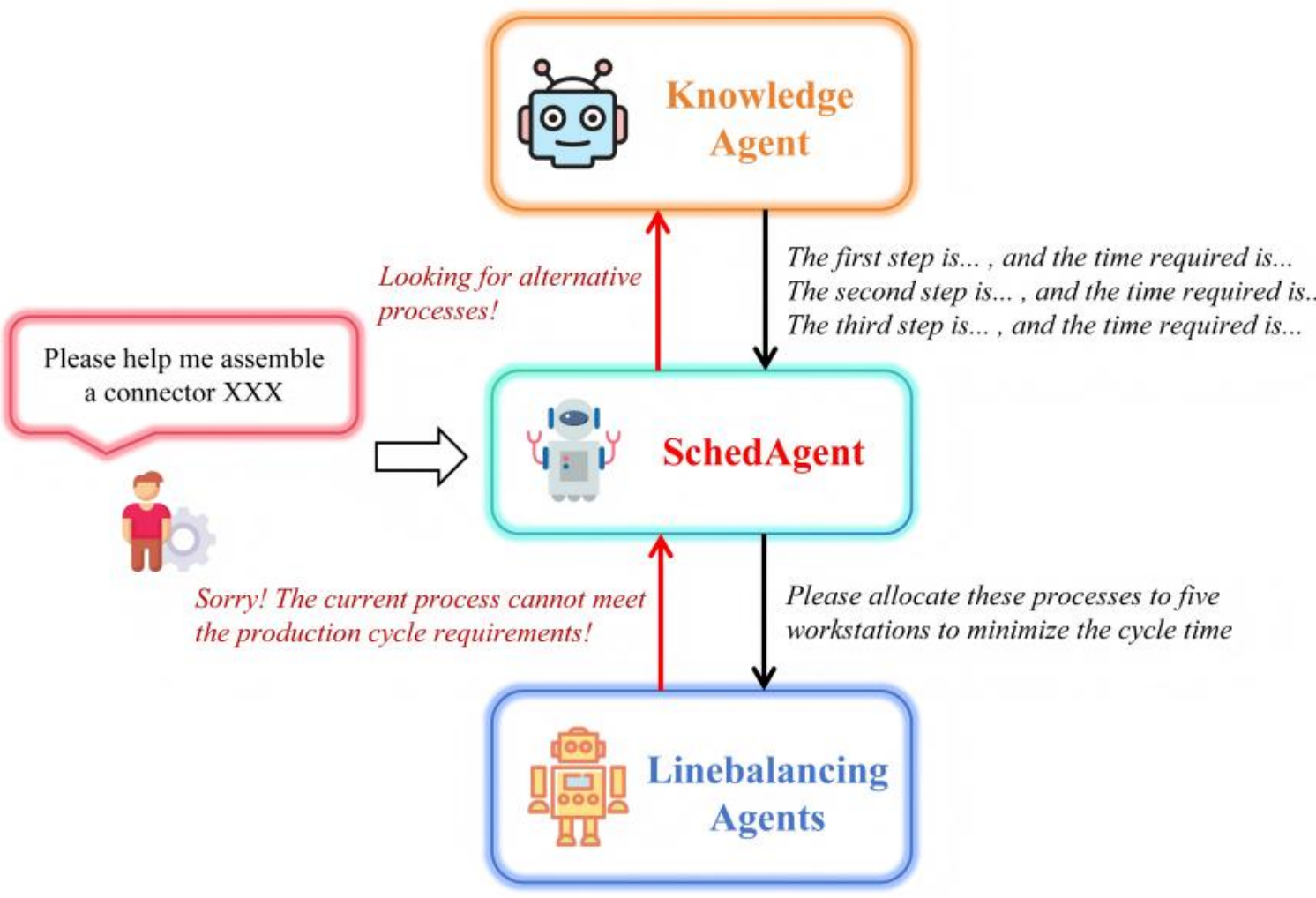

Fig. 2 The dynamic negotiation and coordination mechanism among agents

**SceneGraph:** The scene graph represents production resources and their relational dependencies in a structured graphical format. It delineates the topological connections among resources and exhibits robust scalability, facilitating dynamic updates in response to evolving production tasks. By interrogating the scene graph, the SchedAgent retrieves precise spatial coordinates and resource attributes, thereby enabling optimized allocation and ensuring the conflict-free execution of assembly operations. Leveraging prior environmental mapping research from our laboratory [50], the system supports rapid scene graph synchronization within novel workspace. To ensure seamless integration, the SchedAgent interacts with the scene graph via a serialized API, which transforms graph triplets into high-fidelity semantic contexts for LLM based reasoning.

## 3.3 KnowledgeAgent

In complex industrial assembly environments, effective task planning necessitates the seamless integration of real-time resource scheduling and an extensive repository of structured domain expertise. The KnowledgeAgent functions as the core component for retrieving, synthesizing, and formalizing domain-specific knowledge from industrial knowledge base. It provides the essential logical grounding required for downstream agents to make informed decisions during production planning and assembly execution. Leveraging a RAG architecture with knowledge graphs, the KnowledgeAgent integrates retrieved information fragments to generate coherent, contextually accurate responses.

Upon receiving a query $q_t$ , the KnowledgeAgent executes a multi-stage workflow encompassing context-aware retrieval, semantic reasoning, and knowledge synthesis. It functions as a continuous information provider for the SchedAgent, ensuring that all generated domain insights are contextually relevant and formally actionable for assembly planning. The process of knowledge retrieval and integration is formalized as follows:

$$a_{t+1} = G(R(q_t, K_{base})) \tag{4}$$

Where $a_{t+1}$ represents the response generated by KnowledgeAgent at time $t$ for the requesting agent; $R(\bullet)$ represents the retrieval function that extracts pertinent information from the industrial knowledge base $K_{base}$; $G(\bullet)$ denotes the synthesis and reasoning function that integrates the retrieved fragments. The KnowledgeAgent leverages LLM driven inference to synthesize and reason over these retrieval results, maintaining seamless compatibility with the SchedAgent and LineBalanceAgent. By anchoring high-level reasoning in structured domain data, this workflow suppresses hallucinations and ensures that the planning process remains strictly aligned with industrial specifications. Furthermore, the agent supports continuous updates to its knowledge base, thereby enhancing the system's adaptability and scalability in evolving production environments.

To provide the requisite symbolic grounding for autonomous decision making, we construct a domain-specific industrial knowledge base that serves as the system's authoritative logic foundation. This repository integrates data sources encompassing part specifications, assembly processes, and manufacturing tooling into a formalized KG. By mapping entities and their intricate topological dependencies, the KG establishes a structured representation that facilitates rigorous symbolic reasoning and complex task querying. Based on this foundation, the following details the construction and structural expansion of the knowledge base, ensuring that the agents are provided with the verified industrial logic necessary for coordinated planning.

**Construction of industrial knowledge base:** In this study, process documents and structured prompts are provided to a LLM to build an industrial knowledge graph (Fig. 3). The construction process includes the following steps: First, key entities (such as part models, assembly processes, required time, and process names) and their attributes are extracted from documents using natural language processing methods and LLM, and relationships among entities are identified (e.g., "Process A is the first step to assemble product C901," "Part A is the reference part for Part B," "The next step after Process A is Process B"). Then, entities and relationships are stored in a graph structure, where nodes represent entities and edges represent functional, spatial, or sequential relations. Next, duplicated or redundant information is merged to ensure the graph remains concise and actionable. Finally, to support efficient retrieval for agents, each entity and relationship is assigned indexed key values including attributes, locations, and associated processes allowing agents to quickly access task-related resources.

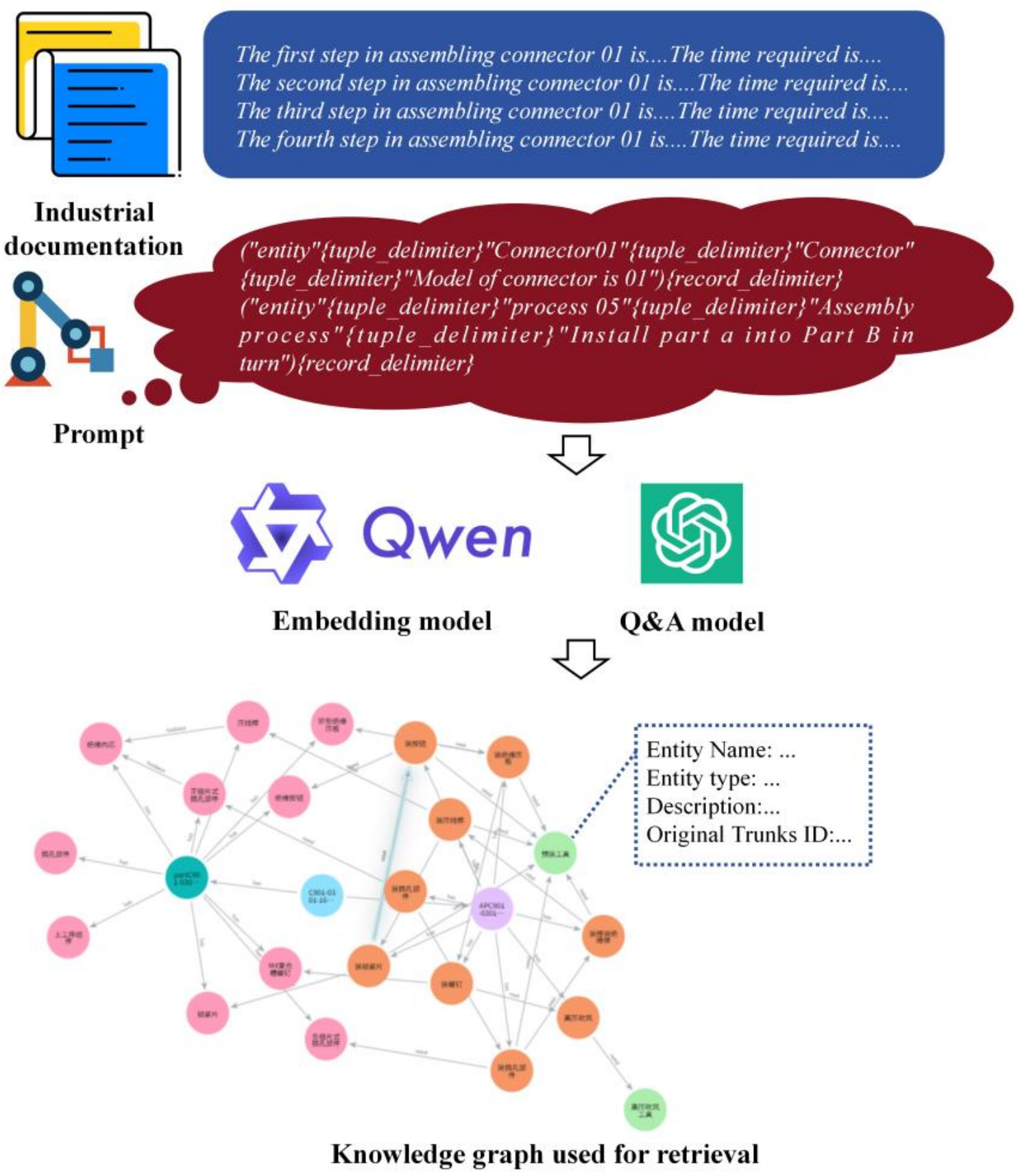

Fig. 3. Construction process of the industrial knowledge graph

**Retrieval of industrial knowledge base**: To enable more effective retrieval from the constructed knowledge base, this study adopts a two-layer retrieval paradigm. This approach captures entities and processes relevant to the query at the semantic level while leveraging the structural information of the knowledge graph to expand the retrieval scope, thus obtaining more comprehensive results. Specifically, for a user query $q$, the system first applies a vector based retrieval function $R(q, K_{base})$ to extract a semantically relevant candidate set from the industrial knowledge base $K_{base}$:

$$D = R(q, K_{base}), D \in K \tag{5}$$

$D$ represents the knowledge fragments such as parts, processes, or tools that directly match the query. Subsequently, to enhance the industrial semantic completeness of the retrieval results, entity and relationship nodes in the set $D$ relationship nodes in the set are expanded by one-hop neighbors to form a local subgraph $G_q$. During the generation stage, the initial retrieval results $D$ are integrated with multi-source knowledge from the expanded subgraph $G_q$:

$$\varphi(q) = P(D, G_q) \tag{6}$$

$P(\bullet)$ represents the information integration and formatting function, which transforms task-related knowledge such as process logic, tool requirements, and part locations into structured context. Finally, the processed information $\varphi(q)$ and the query $q$ are jointly fed into the large language model to generate responses aligned with industrial task constraints. Compared with general retrieval methods, this two-layer retrieval paradigm maintains high efficiency while emphasizing structured modeling of industrial knowledge and constraint-aware information capture, providing solid support for downstream agents in reasoning and decision-making under complex assembly scenarios.

**Dynamic expansion of industrial knowledge base:** To better adapt to changing industrial scenarios, the knowledge base needs to incorporate new process documents, part specifications, or assembly procedures without requiring complete reconstruction upon each update. To achieve this, this study introduces a dynamic expansion mechanism that enables the industrial knowledge base to rapidly absorb new domain knowledge while maintaining overall consistency. Specifically, for newly added process documents $D'$, the system first applies an entity and relationship extraction function $\varphi(\bullet)$ to generate the corresponding knowledge subgraph:

$$D' = (V', E') = \varphi(D') \tag{7}$$

$V'$ represents the newly extracted set of industrial entities (such as parts, processes, and tools), $E'$ denotes the functional dependencies, temporal constraints, and spatial relationships among them. Subsequently, the knowledge base integrates the new subgraph with the existing graph through set merging:

$$\tau_{t+1} = \tau_t \cup V', \ \varepsilon_{t+1} = \varepsilon_t \cup E' \tag{8}$$

$\tau_t$ and $\varepsilon_t$ denote the entity and relationship sets of the knowledge base at time t, $\tau_{t+1}$ and $\varepsilon_{t+1}$ represent the expanded complete knowledge base. This approach allows the system to introduce new knowledge without disrupting the existing structure.

### 3.4 LineBalanceAgent

Within complex industrial assembly production, line balancing is a pivotal process for ensuring optimal production efficiency and resource utilization. To mitigate challenges such as CT imbalances, and resource constraints, this study develops the LineBalanceAgent, which is engineered to dynamically allocate processes and optimize CT for cost minimization and efficiency maximization. The LineBalanceAgent leverages a natural-language-driven reflective optimization mechanism (Fig. 4). By playing different roles, these multiple agents conduct natural language conversations regarding various optimization targets, producing answers that meet all criteria through a multi-round iterative process.

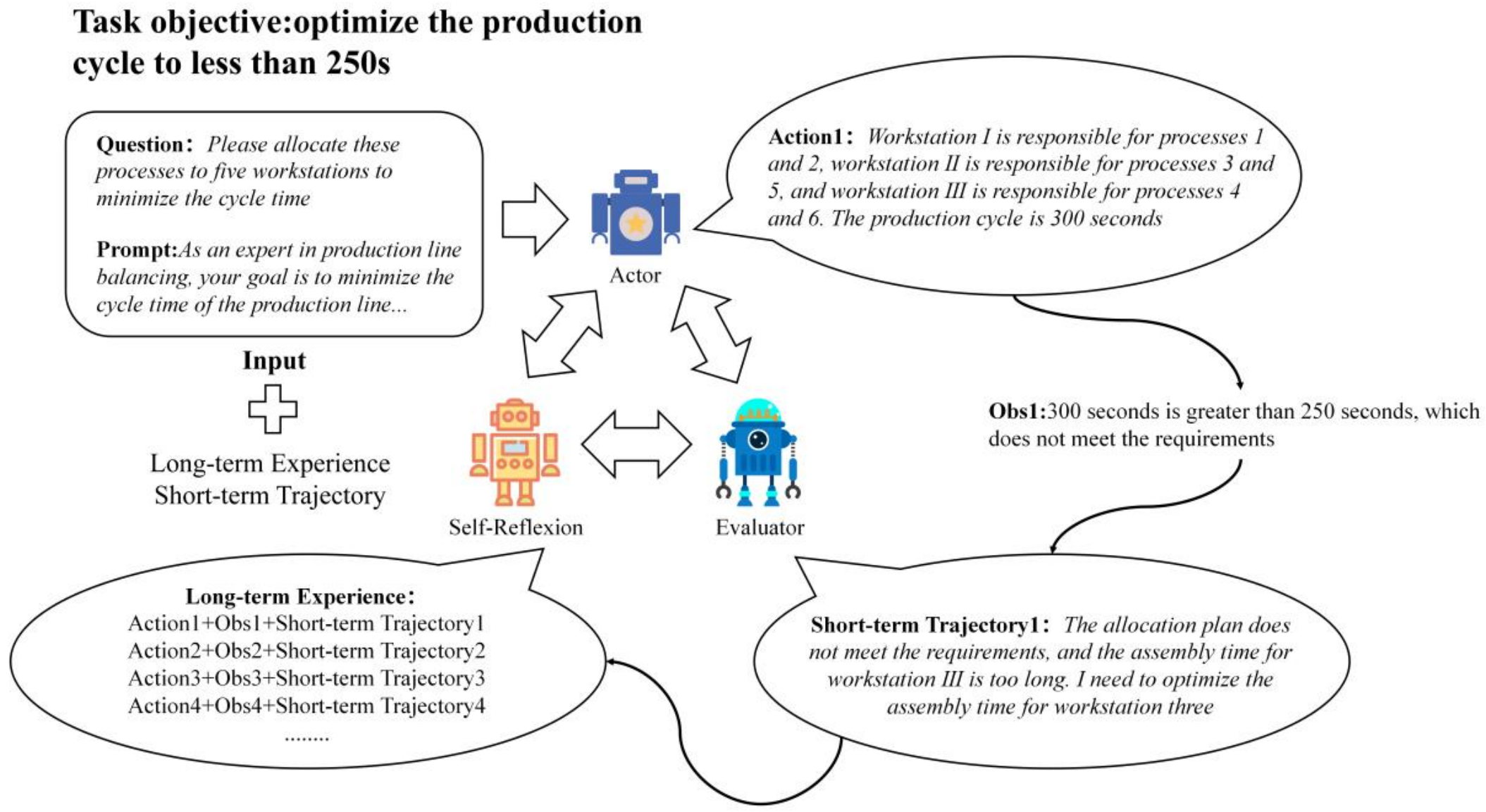

Fig. 4 A self-reflective optimization mechanism driven by natural-language

The LineBalanceAgent consists of three components: the Actor, the Evaluator, and the Self-Reflexion module. The Actor generates actions $a_t$ based on the current context $c_t$, short-term memory $M_t^{short}$ and long-term memory $M_t^{long}$ and subsequently interacts with the production environment $\varepsilon$ which serves as a script to evaluate the compliance of generated schemes with all constraints. The feedback $o_t$ from execution includes information such as workstation load, process execution time, line CT, and resource utilization. The Evaluator performs real-time assessment $e_t$ of the feedback and integrates the results with environmental responses to form short-term memory $M_t^{short} = o_t \cup e_t$ which guides the next decision. Self-Reflexion module combines short-term memory with external constraints $f_t$ to perform self-reflection. It generates reflection outputs $r_t$ and updates long-term memory $M_t^{long} = M_{t-1}^{short} \cup r_t$ to accumulate optimization experience for line balancing. When generating subsequent actions, the Actor refers to both short-term and long-term memories, iterating the above process until achieving a solution that meets line balance objectives such as minimizing the number of workstations or optimizing production CT. This design improves decision quality through multiple rounds of self-reflection while incorporating industrial line constraints and optimization goals into the loop, ensuring practical operability and scalability in real production scenarios. The line balancing process can be expressed as:

$$a_t = \pi(c_t, M_t^{short}, M_t^{long}), o_t = \varepsilon(a_t), M_t^{short} = o_t \cup E(o_t) \quad (9)$$

$$r_t = R(M_t^{short}, f_t), M_t^{long} = M_{t-1}^{long} \cup r_t \quad (10)$$

$$a_{t+1} = \pi(c_{t+1}, M_t^{short}, M_t^{long}) \quad (11)$$

$a_t$ represents the action generated by the Actor at time $t$; $\pi(\cdot)$ denotes the decision function of the Actor; $o_t$ represents the feedback obtained after interacting with the environment, including workstation load, process execution time, and CT information; $E(\cdot)$ denotes the real-time evaluation of the feedback by the Evaluator; $M_t^{short}$ is the short-term memory; $r_t$ represents the reflection result generated by the Self-Reflexion module; $M_t^{long}$ is the long-term memory; $f_t$ refers to external constraints such as tool availability and production line resource limits.

# 4 Experiment

We have performed a comprehensive evaluation of AssemPlanner across three distinct dimensions. First, the KnowledgeAgent was evaluated on industrial knowledge question-answering (QA) tasks. By leveraging various LLM APIs, we analyzed the impact of different model architectures and parameter scales on the framework's performance. Second, we assessed the LineBalanceAgent's performance in production line balancing, comparing its results against baseline methods including traditional search heuristics and reinforcement learning approaches. Finally, the overall task decomposition capability of AssemPlanner was examined. Building upon the foundational work established in [12, 13], the generated subtask sequences were translated into executable robotic code and deployed on a physical pressure reducing valve assembly line. This validation confirms the end-to-end operational feasibility of the framework, from high-level natural-language instructions to physical robotic execution.

## 4.1 Verification of KnowledgeAgent in Industrial Knowledge QA

This section provides a rigorous evaluation of the KnowledgeAgent's performance in industrial knowledge QA tasks. To benchmark the system, a comparative study was conducted using several prominent LLMs, including DeepSeek-R1, Llama 4, Llama 3, Qwen 3, GPT-4, and GPT-3.5. The experiments measure answer accuracy across both single-hop and multi-hop reasoning tasks, investigating how model architecture and scale influence the agent's overall cognitive capabilities.

As illustrated in Table 1, the evaluation framework categorizes tasks based on their logical complexity. Single-hop questions focus on foundational process understanding and applicability judgment. In contrast, multi-hop questions encompass more sophisticated industrial reasoning, such as sequence comparison, sequence linking, requirement querying, and relational mapping. This granular analysis elucidates the performance variations of each model when confronted with the intricate constraints of industrial assembly. To ensure the objectivity and accuracy of the evaluation, a two-stage validation protocol is implemented: the correctness of each response is initially assessed by a LLM, followed by a rigorous secondary verification conducted by three domain experts.

To quantitatively analyze performance, the average accuracy metric $A_m$ is introduced to evaluate the overall performance of model $m$ across all test questions:

$$A_m = \frac{1}{N} \sum_{i=1}^{N} A_{i,m} \times 100\% \tag{12}$$

$A_{i,m}$ denotes the accuracy of model m on question i, and $N$ denotes the total number of questions.

As detailed in Table 1, the experimental results reveal that for single-hop questions, virtually all tested models attained or approached 100% accuracy in overall process understanding. This suggests a foundational competency across modern LLM in mastering basic industrial process knowledge. In terms of process applicability tasks, DeepSeek-R1, Llama 3, and GPT-4 demonstrated superior performance, maintaining near-perfect accuracy. In comparison, Llama 4 and Qwen 3 achieved 94.74% and 97.47%, respectively. GPT-3.5 exhibited a marginal performance deficit compared to its counterparts, reflecting subtle variations in fundamental knowledge extraction and generalization capabilities among the various LLM architectures.

Conversely, multi-hop questions impose more rigorous demands on logical reasoning and multi-step information synthesis. Pronounced performance disparities emerged across specific categories, including sequence comparison, sequence linking, requirement querying, and relational mapping. GPT-4 and DeepSeek-R1 consistently delivered robust, near-perfect accuracy on these multi-hop tasks, demonstrating exceptional proficiency in cross-fragment reasoning and the analysis of intricate process details. Meanwhile, GPT-3.5 showed a substantial performance degradation; its accuracy fell below 80% particularly in requirement query tasks. This decline underscores the inherent limitations of earlier-generation models in executing complex relational reasoning and integrating information across multiple logical steps within an industrial context.

Table 1. Comparison of KnowledgeAgent's accuracy on industrial question answering across different LLMs' API

| **Question Type** | | **Example** | **Number of questions** | **DeepseekR1** | **llama4** | **llama3** | **Qwen3** | **GPT-4** | **GPT-3.5** |
|---|---|---|---|---|---|---|---|---|---|
| Single-hop Questions | Overall Process Understanding | What is the complete assembly process for connector C901? | 19 | 100% | 100% | 100% | 94.74% | 100% | 89.47% |
| | Process Applicability | Which products require the "laser marking" process for assembly? | 38 | 100% | 94.74% | 100% | 97.47% | 100% | 92.11% |
| Multi-hop Questions | Sequence Comparison | What is the first assembly step for connector C901? | 166 | 99.39% | 93.37% | 87.35% | 90.36% | 100% | 92.17% |
| | Sequence Linking | What is the next step after the first assembly process of connector C901? | 166 | 97.96% | 91.57% | 92.77% | 96.99% | 98.8% | 86.75% |
| | Requirement Query | Which components and tools are required for the first assembly step of connector C901? | 74 | 98.65% | 97.39% | 97.39% | 90.54% | 98.65% | 75.68% |
| | Relation Comparison | In step 05 of inserting socket parts during connector C901 assembly, which part serves as the reference component? | 96 | 96.88% | 95.83% | 94.79% | 94.79% | 97.92% | 90.63% |

To provide a more holistic assessment of each model's capability across diverse question categories, this study introduces the Problem Type Weighted Accuracy (PTWA) metric. This metric facilitates a quantitative ranking of the models (as illustrated in Fig. 5) and is formally defined as:

$$P_m = \sum_{t \in T} \omega_t \times A_{m,t} \tag{13}$$

$\omega_t$ denotes the weight of problem type t, which can be set according to practical application needs. In this study, equal importance is assigned to single-hop and multi-hop questions, with both weights set to 0.5. For ease of analysis, two basic evaluation metrics are also defined: Single-Hop Accuracy (SHA) and Multi-Hop Accuracy (MHA).

By leveraging these metrics, the overall efficacy of each model within the KnowledgeAgent framework can be quantitatively benchmarked, providing a rigorous reference for subsequent agent optimization and industrial deployment. The experimental results demonstrate that GPT-4 and DeepSeek-R1 achieved the highest overall performance, followed by Llama 4, Llama 3, and Qwen 3, while GPT-3.5 exhibited the lowest overall efficacy.

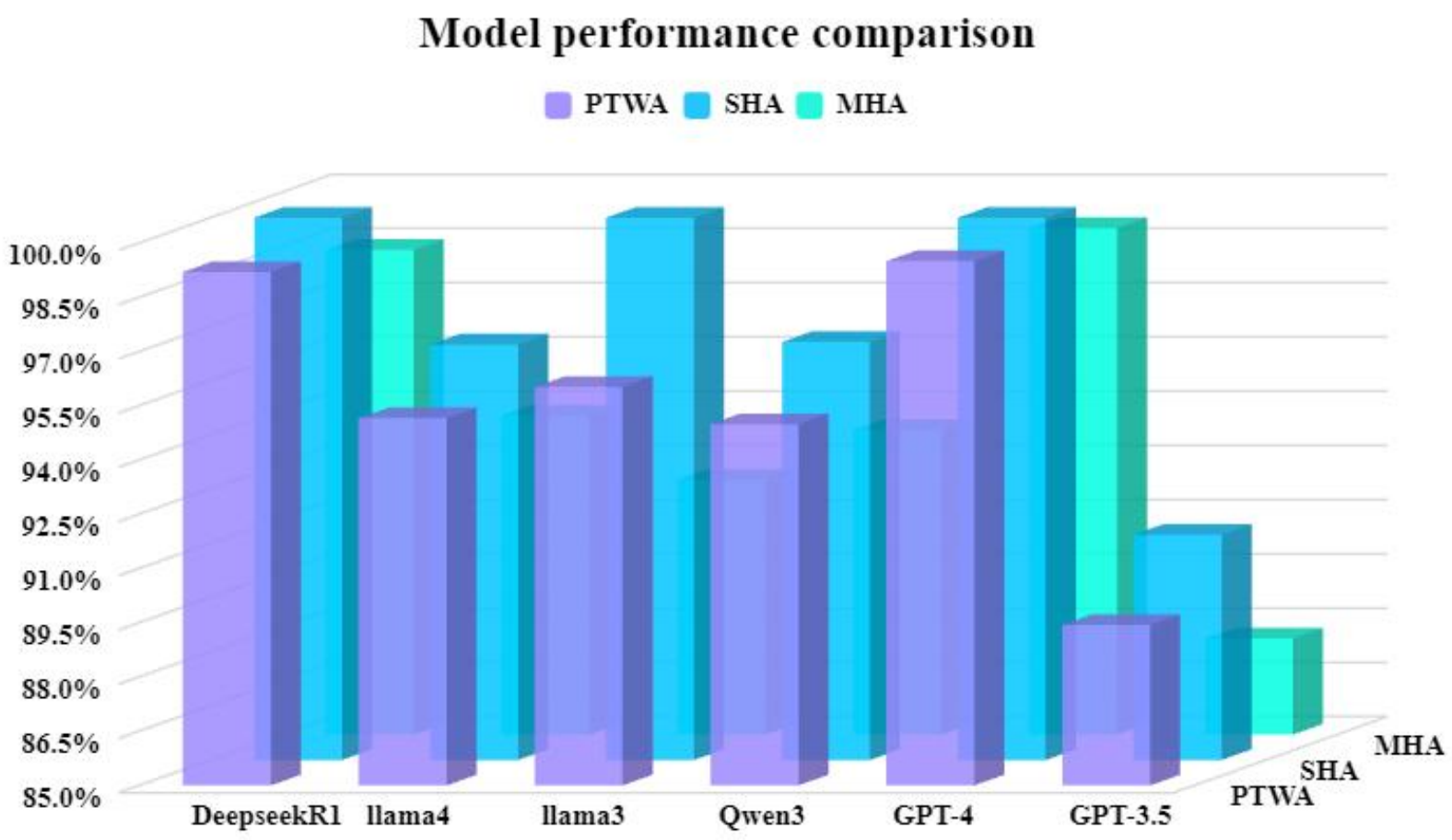

Fig.5 Accuracy Metrics of Models in KnowledgeAgent Industrial QA Task

Addressing the rigorous demands of industrial environments, this study implements a hybrid retrieval strategy that synergizes graph structured data with semantic information. This dual path approach ensures comprehensive and contextually relevant support across diverse question typologies. Experimental results demonstrate that all evaluated models maintain consistent performance in single-hop tasks, underscoring a robust foundational capability in industrial process comprehension and information extraction when supported by structured contextual grounding. However, as task complexity escalates to multi-hop scenarios requiring cross fragment logical reasoning and the integration of multi-step constraints pronounced performance disparities emerge. While GPT-4 and DeepSeek-R1 sustain high precision in these complex reasoning domains, models with lower reasoning capacities exhibit significant accuracy degradation, particularly in requirement querying and sequence linking. These findings emphasize that sophisticated reasoning and strict constraint preservation are equally indispensable for ensuring the high reliability required in industrial grade production planning.

## 4.2 Evaluation of LineBalanceAgent for Optimizing Production Line Balancing

This section provides a systematic evaluation of the LineBalanceAgent in industrial production line balancing tasks. The primary objective is to verify the agent's ability to optimize process allocation and achieve optimal or near-optimal CT while significantly reducing the required iteration counts. To benchmark its performance, the LineBalanceAgent is compared against mainstream scheduling methods: DQN (representing learning based approaches) and IGA, MBO, ABC, and ZOA (representing state-of-art meta-heuristic algorithms commonly utilized in flexible workshop scheduling). Furthermore, to quantify the contribution of the proposed natural-language-driven reflective optimization mechanism, an ablation study was conducted. In this experiment, the full LineBalanceAgent is compared against a baseline utilizing direct GPT-4 reasoning. This comparison isolates the impact of the iterative "Reason – Action – Observe" cycle on optimization quality.

To ensure alignment with real-world manufacturing, a pressure reducing valve assembly task was selected as the evaluation case. This assembly process comprises 12 operations, each characterized by a fixed execution time and strict precedence constraints. The detailed assembly times and precedence relationships for each operation are summarized in Table 2.

Table 2. Assembly time and precedence relationships for the pressure valve assembly

| **Process ID** | **Assembly Time (s)** | **Predecessor Process** |
|---|---|---|
| Process 1 | 124 | / |
| Process 2 | 82 | 1 |
| Process 3 | 71 | 12 |
| Process 4 | 94 | / |
| Process 5 | 72 | 3，4 |
| Process 6 | 67 | 5 |
| Process 7 | 78 | 5 |
| Process 8 | 102 | / |
| Process 9 | 138 | 8 |
| Process 10 | 102 | / |
| Process 11 | 147 | 9，10 |
| Process 12 | 58 | 11 |

To ensure a rigorous and fair comparison, all methods were executed under identical assembly line configurations. This setup included a fixed allocation of five workstations, a consistent distribution of process duration, a target CT constraint of $CT < 250s$ , and standardized tool switching rules. To comprehensively evaluate each method's performance in real world production tasks, five industry standard metrics were defined:

**Cycle Time (CT, s):** The processing time of the bottleneck workstation, representing the production CT. A value below 250s is considered acceptable. The calculation formula is as follows:

$$CT = \max_{\omega \in W} T_{\omega} \tag{14}$$

$T_{\omega}$ denotes the total processing time of all operations assigned to a workstation $\omega$. The production line is considered qualified when the condition is satisfied.

**Line Balancing Rate (Line Balancing Rate, LBR, %):** Indicates the load balance level among workstations which is distributed as:

$$LBR = \frac{\sum_{\omega-1}^{m} t_{\omega}}{m \times CT} \times 100\% \tag{15}$$

$t_{\omega}$ represents the load time of the workstation $\omega$, and $m$ denotes the total number of workstations, a higher value indicates a lower degree of bottleneck.

**Number of Iterations to Converge (NITC):** The number of iterations required to reach a feasible solution, used to measure scheduling response speed;

**Number of Work Units (NWU):** The number of workstations required to complete the task. fewer units indicate higher resource utilization efficiency;

**Tool Change Times (TCT):** The number of tool or fixture changes. fewer changes indicate shorter setup time and improved operational efficiency.

Table 3. Performance comparison of different methods in production line balancing

| Method | NITC | NWU | CT(s) | TCT | LBR (%) |
|---|---|---|---|---|---|
| IGA | 200 | 6 | 234 | 5 | 69.3 |
| DQN | 200 | 5 | 243 | 6 | 93.4 |
| MBO | 10 | 5 | 247 | 6 | 91.9 |
| ABC | 690 | 5 | 247 | 6 | 91.9 |
| ZOA | 161 | 5 | 247 | 6 | 91.9 |
| LineBalanceAgent | 8 | 5 | 247 | 6 | 91.9 |
| GPT-4(w/o reflexion) | 1 | 5 | 249 | 7 | 79.0 |

For accuracy, all experiments were repeated five times, and the results were averaged to produce the final findings.

**Ablation Study:** The ablation experiment reveals the critical role of the reflective optimization mechanism. While the GPT-4 (without reflexion) baseline achieves a feasible solution (CT=249s) in just a single iteration (NITC = 1), it delivers the second lowest Line Balancing Rate (79.0%) and the highest Tool Change Times (TCT=7). In contrast, the full LineBalanceAgent significantly improves the LBR to 91.9% and reduces TCT to 6 within only 8 iterations. This demonstrates that while direct LLM reasoning can quickly find a valid plan, the iterative reflection process is indispensable for refining process allocation and achieving high-quality industrial optimization.

**Comparative Analysis with Meta-Heuristic Algorithms (MBO, ABC, ZOA, and IGA):** The comparative results demonstrate that the LineBalanceAgent achieves a superior balance between optimization efficiency, solution quality, and resource utilization compared to traditional meta-heuristic algorithms. Specifically, while MBO, ABC, and ZOA eventually converge to an identical LBR of 91.9%, the LineBalanceAgent reaches this state with significantly lower computational overhead. With an NITC of 8, the agent outpaces MBO (NITC=10), ZOA (NITC=161), and ABC (NITC=690), achieving a convergence speed faster than the latter. This underscores the efficacy of LLM based semantic reasoning in navigating complex constraint spaces more effectively than Heuristic search algorithm. Furthermore, in comparison with IGA, the proposed agent maintains higher resource efficiency. Although IGA achieves the minimum CT (234s), it does so by increasing the hardware footprint, requiring six workstations (NWU=6) whereas the LineBalanceAgent utilizes only 5. The significantly lower LBR of IGA (69.3%) further suggests that its heuristic mechanism prioritizes local CT reduction at the expense of global workload distribution, whereas the LineBalanceAgent ensures a more balanced and cost-effective production configuration for industrial deployment.

**Comparison with Learning based Approach (DQN):** The comparison with DQN highlights the trade-off between optimization limit and deployment cost. Although DQN achieves the highest LBR (93.4%), it requires 200 iterations to converge. In dynamic industrial environments where line reconfiguration is frequent, the computational and time overhead of DQN's iterative learning process often outweighs its marginal gains in balancing rate.

Beyond quantitative metrics, the LineBalanceAgent offers substantial engineering advantages in terms of deployment flexibility and algorithmic maintenance. Traditional meta-heuristic algorithms typically necessitate expert intensive efforts to design problem specific encoding schemes and to manually tune a multitude of hyper parameters, such as population size, mutation rates, or search coefficients, to ensure convergence. Similarly, DQN requires meticulous reward engineering and complex state space definitions tailored to the specific assembly constraints. In contrast, the proposed agent circumvents these expert dependent processes by operating purely on natural-language instructions. By autonomously parsing industrial constraints and optimization goals from textual descriptions, the LineBalanceAgent achieves high quality optimization without the need for manual parameter tuning or specialized rule setting. This paradigm shift enables the system to maintain an ideal balance between optimization quality (LBR), computational efficiency (NITC), and resource utilization (NWU). Its ability to match the performance of established meta-heuristics while dramatically shortening response times and eliminating manual tuning overhead makes it a highly scalable and practical solution for dynamic, reconfigurable assembly environments.

## 4.3 Experimental Validation of the Integrated AssemPlanner Framework

To systematically evaluate the performance of the AssemPlanner framework in industrial task planning, we conducted comprehensive tests using 19 types of connector assembly tasks. Each task represents a complete product assembly process designed to verify the framework's end-to-end planning and execution capabilities. Through the framework's task decomposition mechanism, these 19 global tasks were decomposed into 419 subtasks. We define a subtask as an atomic semantic unit of the assembly sequence, representing a single, logically indivisible operation.

To evaluate the framework's precision in environmental perception and resource management, each subtask was assigned functional attribute labels based on its operational requirements:

Location Labels (332 steps): Assigned to subtasks that require inferring the coordinates of parts and tools. Based on the requirements identified by the KnowledgeAgent, the framework integrates data from the Scene Graph and feedback from the Line Balancing Agent to reason out the current positions of resources and their designated placement locations.

Object Labels (242 steps): Assigned to subtasks involving resource retrieval, specifically the physical picking of the required parts and tools.

It is important to note that these labels are multi-dimensional rather than mutually exclusive. A single subtask may be labeled with both attributes if it simultaneously demands spatial positioning and entity identification. For example, the subtask "pick the rubber ring tool from Shelf 2 in Room 2" is counted in both categories. This classification logic explains why the sum of labeled steps exceeds the total number of subtasks.

To evaluate the model's generalization and adaptability under different few-shot learning conditions, experiments were conducted in zero-shot (no example), one-shot (single example), and two-shot (dual example) scenarios. To ensure the objectivity and reliability of the evaluation, the correctness of the planning results at each level was manually verified by three independent industrial experts. Each expert assessed the logical feasibility and operational accuracy of the generated plans; A result was deemed correct only when a consensus was reached through a majority voting mechanism. Accuracy was calculated across four levels: Task, Subtask, Location, and Object. The definition of accuracy for each category is as follows, taking Task accuracy as an example:

$$Acc_{task} = \frac{N_{task}^{correct}}{N_{task}^{total}} \tag{16}$$

$N_{task}^{correct}$ denotes the number of tasks with completely correct end-to-end planning. $N_{task}^{total}$ represents the total number of tasks. Other level metrics are calculated using the corresponding number of correctly matched steps and total step count. The experimental results are shown in Table 4:

Table 4. Accuracy of AssemPlanner across different task levels under varying shot conditions

| | **Task** | **Subtask** | **Location** | **Object** |
|---|---|---|---|---|
| **Zero-shot** | 26.32% | 85.89% | 85.13% | 79.57% |
| **One-shot** | 63.16% | 93.81% | 92.49% | 89.71% |
| **Two-shot** | 68.42% | 97.22% | 96.46% | 86.94% |

To account for the inherent stochasticity of LLMs and ensure the reliability of the results, each experiment was repeated five times under identical conditions. The results reported in this section represent the mean values across these five independent runs, and the accuracy levels across the five trials were highly consistent with minimal differences. This confirms the stability and statistical significance of the AssemPlanner framework.

In the zero-shot setting, AssemPlanner achieved relatively low accuracy at the Task level (26.32%), but exceeded 79% accuracy at the Subtask, Location, and Object levels, indicating strong task decomposition and understanding capabilities of environment and object even without prior examples. With the introduction of one-shot examples, all metrics improved significantly, with the Task level showing the largest increase (+36.84%), demonstrating that AssemPlanner can quickly adapt to new assembly tasks after a single demonstration. In the two-shot scenario, task accuracy increased slightly (+5.26%), while subtask and the location and object recognition accuracy approached 100%, suggesting performance saturation. A web interface (Fig. 6) was developed to visualize the process of AssemPlanner performing task planning after receiving natural-language instructions.

To evaluate the effectiveness of each core component within the AssemPlanner framework, we conducted ablation experiments by systematically removing the KnowledgeAgent, the Scene Graph, and the LineBalanceAgent (tested under a two-shot setting). Our results demonstrate that these modules are indispensable for a flexible assembly task planning system, as illustrated in figure 6:

Configuration without KnowledgeAgent: The framework's performance across all evaluation metrics include Task, Subtask, and Location drops to zero. In the absence of the KnowledgeAgent, the system suffers from total semantic hallucinations, as it is unable to retrieve domain specific connector assembly processes.

Configuration without Scene Graph: The agent loses its spatial grounding. While the system retains the ability to generate logical assembly sequences, it becomes incapable of configuring the specific tools and parts required for each assembly unit. Consequently, the accuracy for all tasks related to Location and Object labels drops to zero.

Table 5. Ablation Experiment on AssmPlaner in the case of two-shot

| | **Task** | **Subtask** | **Location** | **Object** |
|---|---|---|---|---|
| **AssmPlanner w/o KnowledgeAgent** | 0.00% | 0.00% | 0.00% | 0.00% |
| **AssmPlanner w/o Scene graph** | 0.00% | 21.99% | 0.00% | 0.00% |
| **AssmPlanner** | 68.42% | 97.22% | 96.46% | 86.94% |

Configuration without LineBalanceAgent: While the agent can still generate basic operational sequences, it fails to account for workstation allocation constraints and production CT limits. Consequently, the resulting subtask list becomes practically meaningless for real world assembly, as it cannot satisfy the rigorous synchronization requirements of a production line. Without the LineBalanceAgent, the framework is unable to align task distribution with the physical boundaries of workstations or the temporal demands of production CT, rendering the entire plan invalid in multi-constraint industrial scenarios.

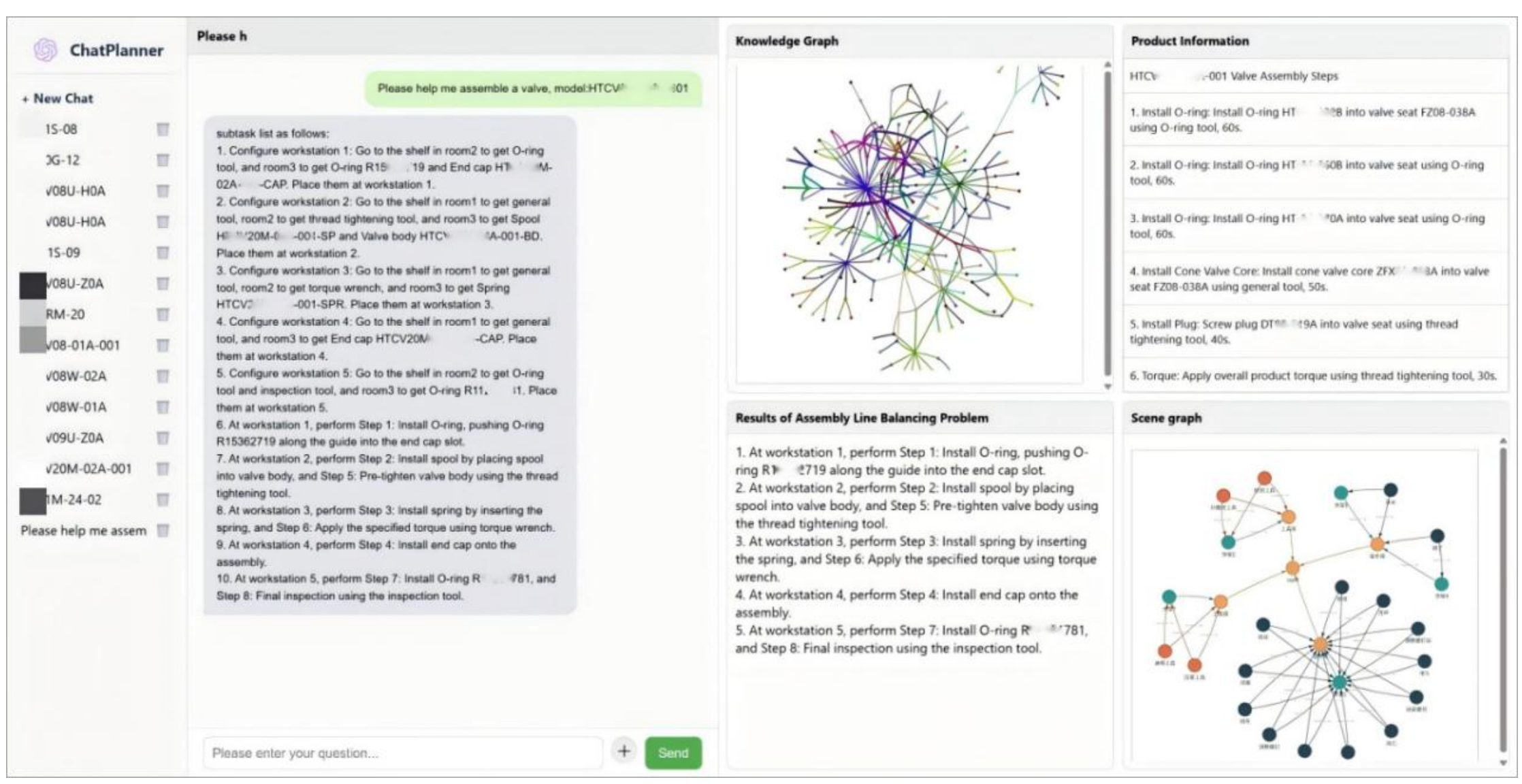

Fig. 6 Example of task-level planning visualization in AssemPlanner framework(Certain sensitive product information has been masked in the figure due to confidentiality requirements. This does not affect the analysis or conclusions presented in this study.)

Experimental results indicate that AssemPlanner maintains robust performance in task decomposition, spatial localization, and resource identification, even in zero-shot settings. This resilience is primarily attributed to its multi-agent collaborative architecture and the stability of the knowledge enhanced reasoning mechanism when handling fine-grained tasks. Nevertheless, for comprehensive planning tasks which necessitate the orchestration of multiple subtask sequences while adhering to intricate process dependencies and resource constraints the system is more susceptible to sequence deviations and resource mismatches in the absence of contextual examples. The introduction of few-shot examples significantly bolsters the accuracy of high-level planning, particularly in sequence organization, dependency mapping, and cross-step information synthesis, thereby demonstrating the framework's potent adaptation capability. Notably, performance on fine-grained tasks reaches a saturation point after only a limited number of examples, suggesting that the system can achieve rapid cross-product and cross-line migration with minimal sampling overhead. From an industrial perspective, these findings imply that in multi-variety and small-batch manufacturing scenarios, AssemPlanner can drastically reduce the commissioning and debugging lead time required for production line reconfiguration while maintaining high precision and minimizing the necessity for human intervention.

### 4.4 Case Study: End-to-End Assembly of a Pressure Reducing Valve

To validate the practical deployability and reliability of the AssemPlanner framework in industrial settings, we conducted a comprehensive case study on a pressure reducing valve assembly, which involves 12 distinct components and 12 operational steps. This experiment encompasses the entire functional chain from parsing natural-language instructions to executing low-level control on a flexible assembly line. To rigorously assess the framework's stability, we conducted multiple experimental trials. The results across all iterations remained highly consistent and aligned with our previous established benchmarks [13]. This high degree of reproducibility demonstrates that by grounding the reasoning process in a domain specific Knowledge Graph and a structured multi-agent coordination mechanism, AssemPlanner effectively mitigates the hallucination typically associated with LLM. Consequently, the framework ensures the generation of deterministic and executable assembly sequences, fulfilling the stringent accuracy requirements of physical production environments.

Upon receiving a natural-language instruction, such as 'assemble a pressure-reducing valve,' the SchedAgent initiates a reasoning process governed by the ReAct paradigm (see Section 3.2). The agent iteratively queries the KnowledgeAgent (detailed in Section 3.3) to retrieve domain-specific expertise, thereby establishing a complete sequence of assembly operations and their estimated execution times. As illustrated in Fig. 6, the knowledge graph for this valve provides the necessary ontological data for all components. Subsequently, the LineBalanceAgent is invoked to distribute these operations across various assembly units, optimizing line balance under predefined CT constraints. This hierarchical reasoning process culminates in a detailed Subtask List, which specifies the execution order, component parameters, and resource requirements. The final task planning and workstation allocation results are visualized in Fig. 7.

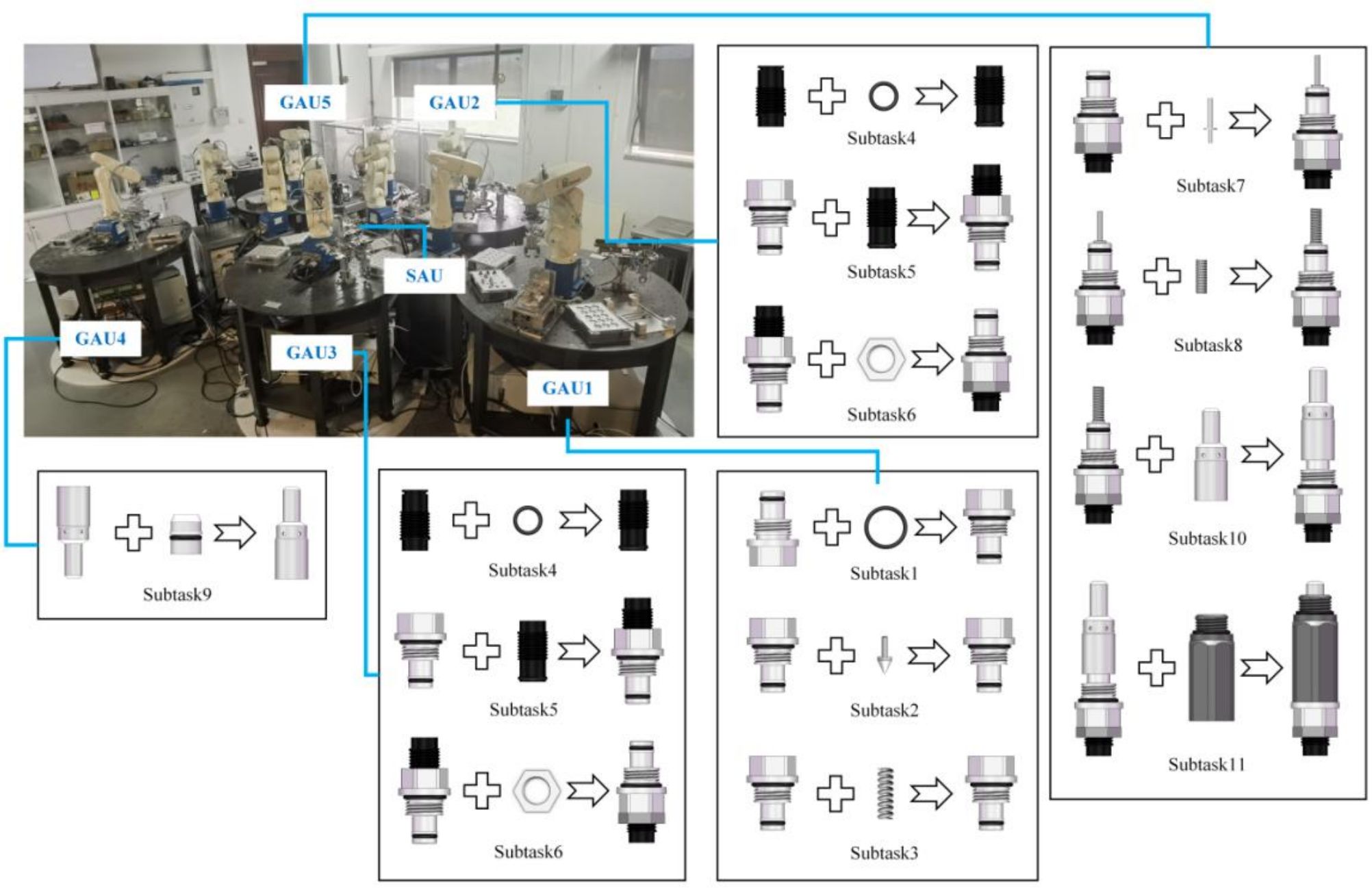

Fig. 7 High-level task planning results.

The Subtask List is then passed to the Low-level Task Planner, which executes the following two stages:

Stage 1: Skill Mapping. Based on predefined Cypher rules, each subtask is decomposed into Start Step, Operation Step, and End Step, and then mapped to specific assembly skills (e.g., Pick and Place, Press Fit, Screw, Rubber Ring Assembly, etc.). This stage eliminates redundant steps such as repeated Start/End actions within the same reference part and generates an optimized sequence of executable skills.

Stage 2: Skill Parameter Completion. For each skill, key parameters such as assembly position, assembly depth, and tool specifications are inferred using rules defined in the product information, scene graph, and knowledge graph.

After skill mapping and parameter completion, the low-level planner converts each skill into corresponding assembly language instruction segments such as mov → grasp → mov → loosen, to generate a complete execution program.

These programs are executed on the Reconfigurable Flexible Assembly System (RFAS) (Fig. 7). The RFAS consists of multiple configurable units (GAU, TU, SAU) equipped with specialized fixtures and tools designed for the pressure reducing valve product family. Based on the task sequence generated by high-level planning, the system automatically selects appropriate assembly units and workstation resources, executing all subtasks according to the balanced optimization strategy.

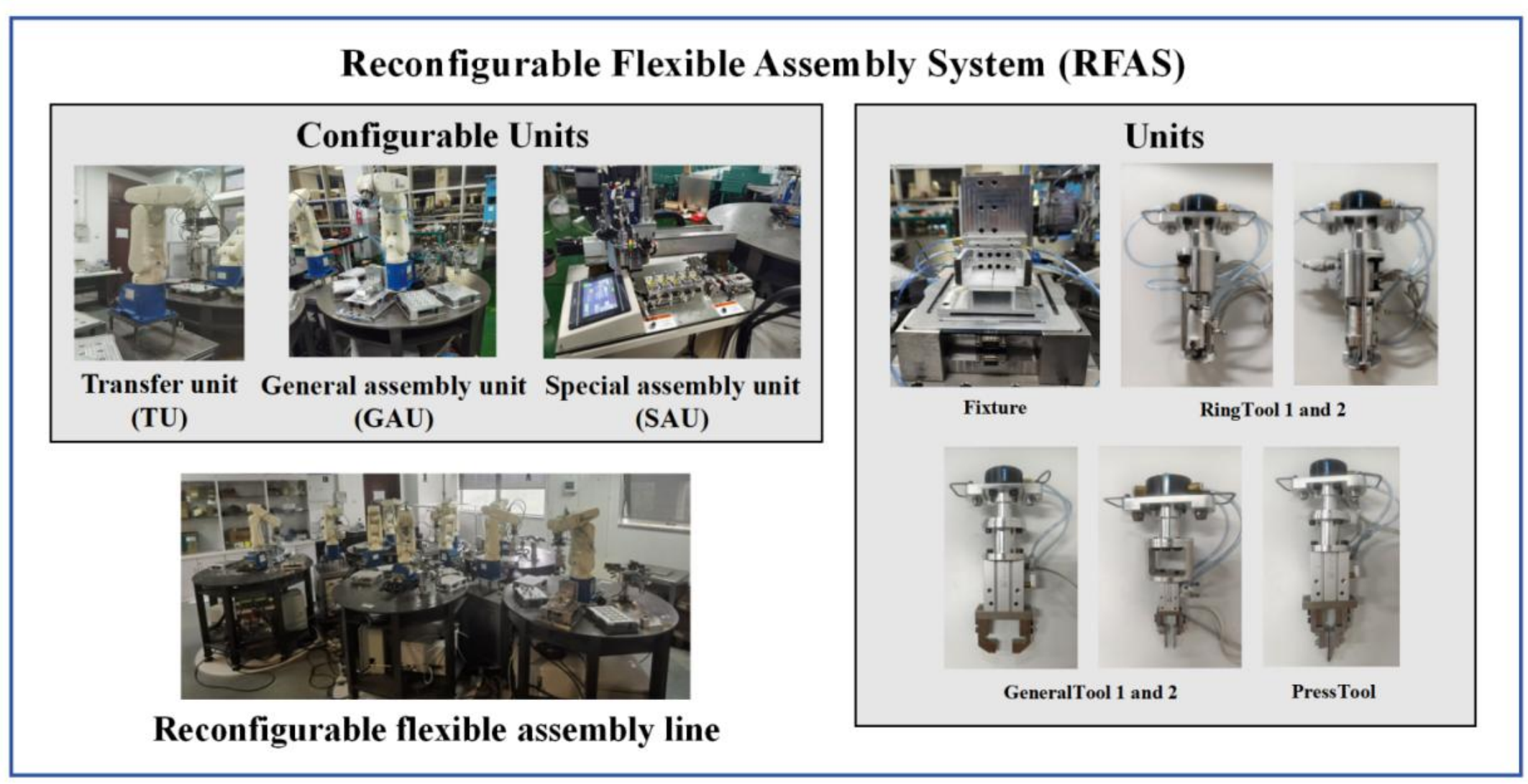

Fig. 7 Reconfigurable flexible assembly system

Taking the pressure reducing valve case as an example, the complete process includes multiple assembly operations such as core installation, rubber ring fitting, tightening, and press fitting. Under the instruction control of AssemPlanner, robotic arms on the production line achieve full automation from part fetching and assembly to tool switching (see Fig. 8). Test results show that AssemPlanner, without additional human intervention, can accurately map high-level natural-language tasks into executable low-level assembly programs and execute them stably on the physical production line, demonstrating the framework's deployability and general applicability within the EIIR closed-loop system.

While this study exemplifies an end-to-end closed loop using a pressure reducing valve as a case study, the architectural core of AssemPlanner lies in the profound decoupling of the planning engine from the domain specific knowledge. This decoupling mechanism endows the system with robust generalization capabilities. When encountering new products, such as connectors, the underlying logical models specifically the ReAct reasoning paradigm of the SchedAgent and the optimization algorithms of the LineBalanceAgent require no parameter fine-tuning or retraining. Users need only update the knowledge base within the KnowledgeAgent. This allows the system to automatically execute skill mapping and parameter completion logic. Consequently, AssemPlanner facilitates zero-code cross-domain migration from specific use cases to broader precision assembly tasks in discrete manufacturing, satisfying the stringent requirements for rapid changeover in RFAS from both theoretical and engineering perspectives

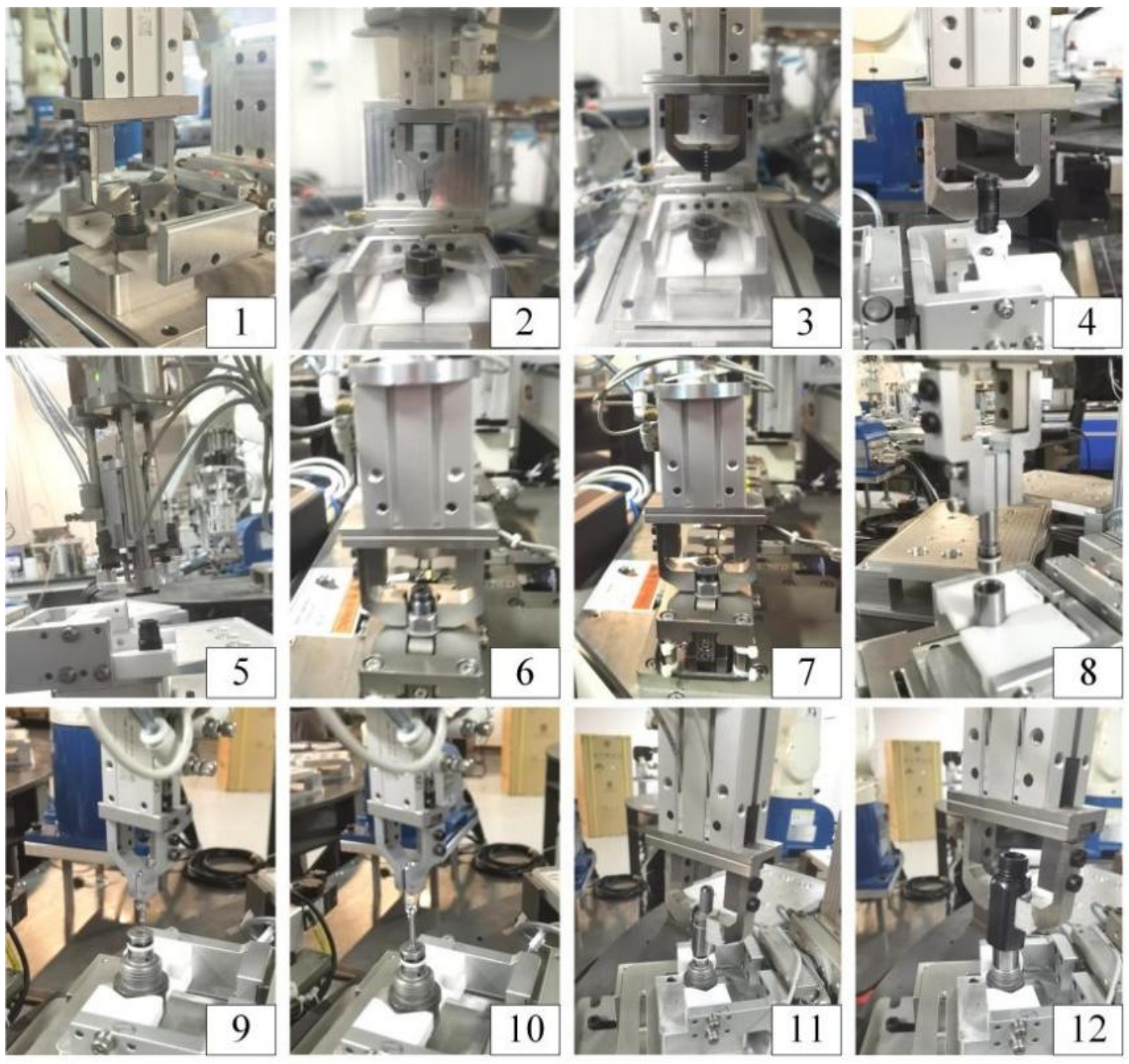

Fig. 8 Actual assembly process

## 5 Conclusion and future work

This paper introduces AssemPlanner, a multi-agent based assembly task planning framework designed as the central planning and scheduling engine of the EIIR system. By seamlessly integrating semantic comprehension, knowledge reasoning, and action execution, the framework effectively bridges the gap between natural-language inputs and production line assembly execution. It facilitates the rapid transformation of natural-language industrial instructions into executable action sequences, supporting agile deployment across diverse products and production environments. System level experimental results demonstrate that the KnowledgeAgent attains high precision in both single-hop and multi-hop reasoning, with GPT-4 and DeepSeek-R1 achieving near perfect scores on the PTWA benchmark. Regarding flexible line balancing, the LineBalanceAgent achieves performance parity with established methods such as Deep Q-Learning and meta-heuristic algorithms. Crucially, AssemPlanner obviates the need for expert level parameter tuning and reward engineering, enabling rapid task adaptation through intuitive textual prompts. Furthermore, the framework showed substantial accuracy improvements in one-shot and two-shot settings, with Subtask, Location, and Object accuracy exceeding 96% in the latter. Finally, a real world pressure reducing valve assembly case validated the end-to-end closed-loop capability of AssemPlanner, confirming its robustness and feasibility for physical production line deployment.

The deployment of AssemPlanner offers three strategic advantages for industrial management:

**Operational Agility:** The framework enables rapid adaptation to small batch, customized assembly. By utilizing natural-language instructions rather than manual programming, managers can significantly reduce production line reconfiguration time, enhancing responsiveness to dynamic market demands.

**Reduced Deployment Cost:** It eliminates the need for expert level parameter tuning and complex reward function design. This significantly lowers the technical debt and maintenance costs typically associated with maintaining sophisticated AI scheduling models.

**Workforce Empowerment:** AssemPlanner lowers the technical threshold for system operators. By allowing non-expert staff to manage complex assembly logic through intuitive interaction, enterprises can mitigate the risks posed by the shortage of specialized domain-specific programmers.

Nevertheless, this study still has several noteworthy limitations and directions for improvement. Due to performance variations among LLMs, different LLM exhibit noticeable differences in complex logical reasoning accuracy. Future work can explore model integration and domain-adaptive fine-tuning to enhance the adaptability of lower cost models. The real time updating capability of the knowledge base also requires improvement; Although incremental updates are supported, handling unstructured and heterogeneous data formats still needs refinement. Integrating automatic knowledge extraction and semantic fusion could enable near real-time updates. Regarding cross-domain generality, although the framework has been validated across various assembly tasks, further empirical studies and benchmark tests are needed in highly diverse industries such as automotive, aerospace, and electronics manufacturing to evaluate its generalization performance.

Overall, AssemPlanner achieves effective implementation of the task planning module within the flexible assembly system. Through multi-agent collaboration, knowledge enhancement, and natural-language driven optimization mechanisms, it significantly improves the accuracy, adaptability, and transferability of industrial task planning. With ongoing advances in perception fusion, reasoning enhancement, optimization strategy evolution, and accelerated knowledge updating, AssemPlanner is expected to evolve into a general industrial embodied intelligence hub providing smarter, more robust, efficient, and scalable solutions for intelligent manufacturing.